\documentclass{article}

\usepackage{microtype}
\usepackage{graphicx}
\usepackage{amssymb}
\usepackage{subfigure}
\usepackage{booktabs} 

\usepackage{hyperref}

\usepackage{array,graphicx}
\usepackage{booktabs}
\usepackage{pifont}

\usepackage[accepted]{icml2018}
\usepackage{mathtools}

\newcommand\Mycomb[2][^n]{\prescript{#1\mkern-0.5mu}{}C_{#2}}

\icmltitlerunning{Using Mode Connectivity for Loss Landscape Analysis}

\begin{document}

\twocolumn[
\icmltitle{Using Mode Connectivity for Loss Landscape Analysis \\}



\icmlsetsymbol{equal}{*}

\begin{icmlauthorlist}
\icmlauthor{Akhilesh Gotmare}{to}
\icmlauthor{Nitish Shirish Keskar}{to}
\icmlauthor{Caiming Xiong}{to}
\icmlauthor{Richard Socher}{to}
\end{icmlauthorlist}

\icmlaffiliation{to}{Salesforce Research, Palo Alto, USA}

\icmlcorrespondingauthor{Nitish Shirish Keskar}{nkeskar@salesfoce.com}

\icmlkeywords{Machine Learning, ICML}

\vskip 0.3in
]



\printAffiliationsAndNotice{}  

\begin{abstract}

Mode connectivity is a recently introduced framework that empirically establishes the connectedness of minima by finding a high accuracy curve between two independently trained models. To investigate the limits of this setup, we examine the efficacy of this technique in extreme cases where the input models are trained or initialized differently. We find that the procedure is resilient to such changes. Given this finding, we propose using the framework for analyzing loss surfaces and training trajectories more generally, and in this direction, study SGD with cosine annealing and restarts (SGDR). We report that while SGDR moves over barriers in its trajectory, propositions claiming that it converges to and escapes from multiple local minima are not substantiated by our empirical results. 


\end{abstract}

\section{Introduction and Related Work}
\label{submission}

Training neural networks involves optimizing a non-convex objective function with gradient-based methods. Recent work focused on understanding the loss surface of neural networks and the trajectories traced by optimizers like stochastic gradient descent (SGD) and its adaptive variants, including Adam \cite{kingma2014adam}, Adagrad \cite{duchi2011adaptive}, and RMSProp \cite{tieleman2012lecture}. 

\citet{garipov2018loss} introduce a framework to obtain a low loss (or high accuracy, in the case of classification) curve of simple form, such as a piecewise linear curve, that connects optima (modes of the loss function) found independently. This observation suggests that, unlike several empirical results claiming that minima are isolated or have barriers between them, these points, in fact, are connected, at the same loss function depth, via a simple piecewise linear curve. \citet{draxler2018essentially} independently report the same observation for neural network loss landscapes, and claim that this is suggestive of the resilience of neural networks to perturbations in model parameters.

In this work, we present two novel results: first, we evaluate the resilience of the mode connectivity phenomenon by using the proposed procedure to connect optima found via different training schemes, and then proceed to use it as a tool to make observations on the optimization trajectory of SGD with cosine-annealing (SGDR) \cite{loshchilov2016sgdr}. We study this heuristic in particular given its superior empirical performance on many tasks, see e.g., \cite{coleman2017dawnbench}, and also the lack of theoretical motivation explaining why it works. 

We begin by briefly describing the mode connectivity procedure in Section \ref{modeconnprocedure}, and the SGDR strategy in Section \ref{sgdrtrajectory}. In Section \ref{resilience}, we present a short motivation, experimental details and results for testing the mode connectivity framework's resilience. Section \ref{investigative} involves the experiments and analysis of the loss surface and SGDR trajectory.


\section{Mode Connectivity Procedure}
\label{modeconnprocedure}
Let $w_a \in \mathbb{R}^{D}$ and $w_b \in \mathbb{R}^{D}$ be two modes (optimal sets of neural network parameters) in the $D$-dimensional parameter space obtained using independent training runs that both optimize a given loss function $\mathcal{L}(w)$ (like the cross-entropy loss). We represent a curve connecting $w_a$ and $w_b$ by 
$\phi_{\theta}(t) : [0,1] \rightarrow \mathbb{R}^{D}$, such that $\phi_{\theta}(0) = w_a$ and $\phi_{\theta}(1) = w_b$. To find a low loss path, we find the set of parameters $\theta \in \mathbb{R}^{D}$ that minimizes the following loss:
$$\ell(\theta) = \int_{0}^{1} \mathcal{L}(\phi_{\theta}(t))dt = \mathbb{E}_{t \sim U(0,1)} \mathcal{L}(\phi_{\theta}(t))$$
where $U(0,1)$ is the uniform distribution in the interval $[0,1]$.

\begin{figure}[ht]
\begin{center}
\centerline{\includegraphics[scale=0.30]{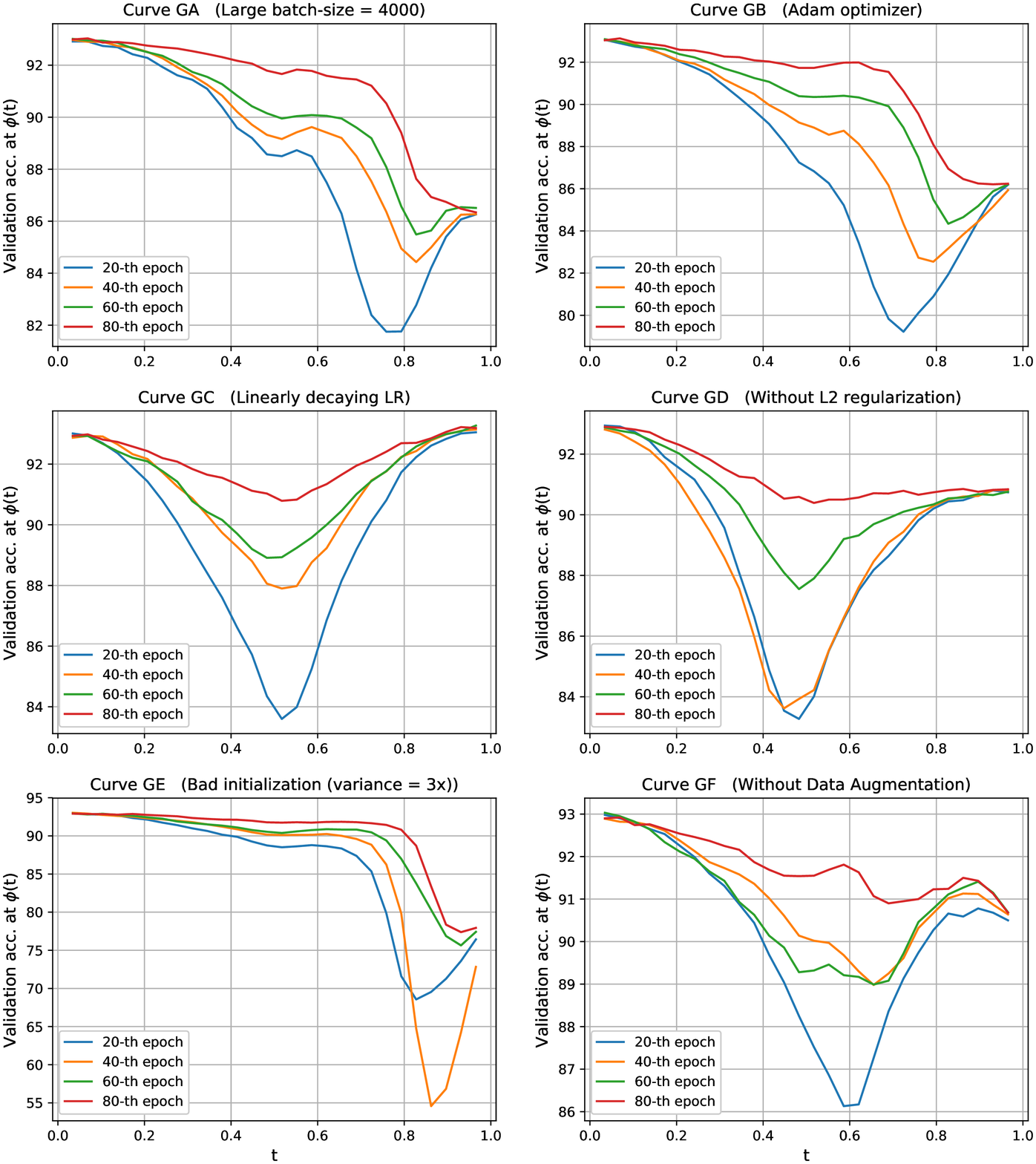}}
\caption{Validation accuracy corresponding to models on the following 6 different curves - curve $GA$ represents curve connecting mode $G$ (one found with all the default hyperparameters) and mode $A$ (found using large batch size), similarly, curve $GB$ connects mode $G$ and mode $B$ (found using Adam), curve $GC$ connects to mode $C$ (found using linearly decaying LR), curve $GD$ to mode $D$ (found with much less L2 regularization), curve $GE$ to mode $E$ (found using a poor initialization), and curve $GF$ to mode $F$ (found without using data augmentation). $t=0$ corresponds to mode $G$ for all of the plots.}
\label{icml-historical}
\end{center}
\vskip -0.2in
\vspace{-2mm}
\end{figure}

To optimize $\ell(\theta)$ for $\theta$, we first need to chose a parametric form for $\phi_{\theta}(t)$. One of the forms proposed by \citet{garipov2018loss} is a polygonal chain with a single bend at $\theta$ as follows 
\[   
{\phi_{\theta}(t)} = 
     \begin{cases}
       2 (t {\theta} + (0.5 - t) w_a), &\quad\text{if } 0 \leq t \leq 0.5 \\
       2 ((t - 0.5)w_b + (1 - t){\theta}) &\quad\text{if } 0.5 < t \leq 1 \\
     \end{cases}
\]
To minimize $\ell(\theta)$, we sample $t \sim U[0,1]$ at each iteration and use the quantity $\nabla_{\theta}\mathcal{L}(\phi_{\theta}(t))$ as an estimate for the true gradient $\nabla_{\theta}\ell(\theta)$ to perform updates on $\theta$ (using SGD). We initialize $\theta$ with $\frac{1}{2}(w_a + w_b)$. Note that in expectation over the uniformly distributed $t$, this computationally cheap estimate is equal to the true gradient 
$$\mathbb{E}_{t \sim U[0,1]}  \nabla_{\theta}\mathcal{L}(\phi_{\theta}(t))  = \nabla_{\theta} \mathbb{E}_{t \sim U[0,1]}  \mathcal{L}(\phi_{\theta}(t)) = \nabla_{\theta}(\ell(\theta)) $$

\section{Resilience of Mode Connectivity}
\label{resilience}
To demonstrate that the curve-finding approach described in Section \ref{modeconnprocedure} works in practice, \citet{garipov2018loss} use two optima found using different initializations but a common training scheme which we detail below. We explore the limits of this procedure by connecting optima obtained from different training strategies. In particular, we experiment with different initializations, optimizers, data augmentation choices, and hyperparameter settings including regularization, training batch sizes and learning rate schemes. 

Conventional wisdom suggests that these different training schemes will converge to different regions in the parameter space that are isolated from each other. Having a high accuracy connection between these pairs would seem counterintuitive. Particularly for the large batch training case, previous works \cite{hochreiter1997flat,keskar2016large} have empirically established that small-batch training leads to wider minima and large-batch training leads to sharper minima. Likewise, with respect to models found using different optimizers, \citet{heusel2017gans} argues that Adam also prefers wide optima and \citet{wilson2017marginal} show that adaptive methods like Adam lead to drastically different solutions from SGD. 
Similarly, in the context of importance of initialization, \citet{goodfellow2016deep} show that the scale of the distribution used for initialization has a large impact on both the outcome of the optimization algorithm and the ability of the network to generalize. Lastly, we know that $L$-2 regularization or weight decay drives the parameters closer to $0$, while a smaller $L$-2 penalty would have a lesser effect of this kind and thus would allow the optimization path to explore models farther from the origin. 
\begin{figure*}[ht]
\centering{\includegraphics[ width = \textwidth]{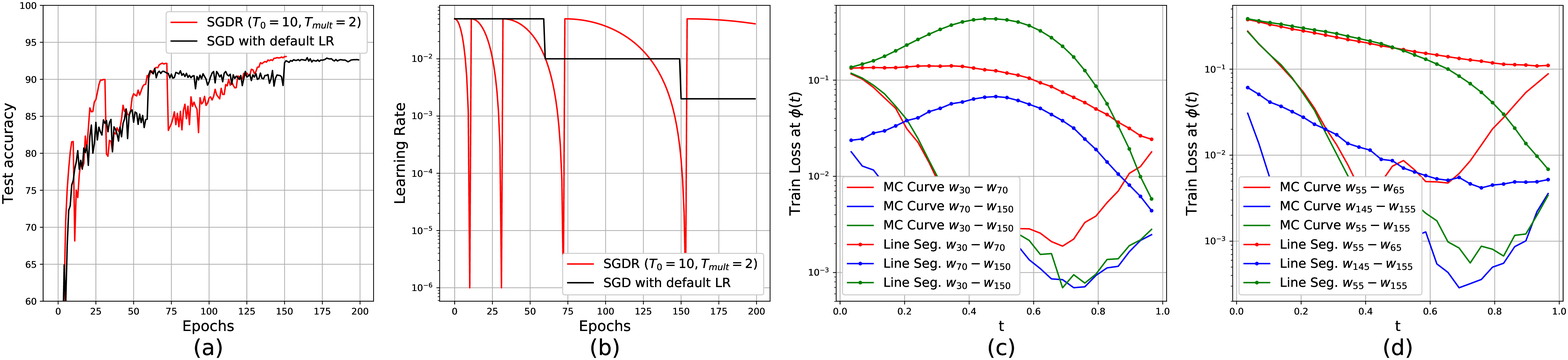}}
\caption{(a) Validation accuracy of a VGG16 model trained on CIFAR10 using SGDR with warm restarts simulated every $T_0 = 10$ epochs and doubling the periods $T_i$ at every new warm restart ($T_{mult} = 2$). (b) SGDR and SGD learning rate schemes. (c) Cross-entropy training loss on the curve found through Mode Connectivity (MC Curve) and on the line segment (Line Seg.) joining modes $w_{30}$ (model corresponding to parameters at the $30$-th epoch of \textbf{SGDR}) and $w_{70}$, $w_{70}$ and $w_{150}$, $w_{30}$ and $w_{150}$. (d) Cross-entropy training loss on the curve found through Mode Connectivity (MC Curve) and on the line segment (Line Seg.) joining modes $w_{55}$ (model corresponding to parameters at the $55$-th epoch of \textbf{SGD with step decay LR scheme}) and $w_{65}$, $w_{145}$ and $w_{155}$, $w_{55}$ and $w_{155}$.
}
\label{sgdr_sgd_modeconnfig}
\vskip -0.2in
\end{figure*}

For obtaining the reference model (named mode $G$), we train the VGG16 model architecture \cite{simonyan2014very} using CIFAR10 training data \cite{krizhevsky2014cifar} for $200$ epochs with SGD. The learning rate is initialized to $0.05$ and scaled down by a factor of $5$ at epochs $\{60,120,160\}$ (step decay). We use a training batch size of 100, momentum of 0.9, and a weight decay of $0.0005$. Elements of the weight vector corresponding to a neuron are initialized randomly from the normal distribution  $\mathcal{N}(0,\,\sqrt[]{2/n})$ where $n$ is the number of inputs to the neuron. We also use data augmentation by random cropping of input images.

We build 6 variants of the reference mode $G$ as follows: we obtain mode $A$ using a training batch size of $4000$, mode $B$ by using the Adam optimizer instead of SGD (and hence also a different learning rate), mode $C$ with a linearly decaying LR scheme instead of the step decay scheme used in mode $G$, mode $D$ using a smaller weight decay of $5 \times 10^{-6}$, mode $E$ by increasing the variance of the initialization distribution to $3 \times \sqrt[]{2/n}$ and mode $F$ using no data augmentation.

Note that for this set of modes $\{A,B,C,D,E,F\}$ all the other hyper-parameters and training settings except the ones mentioned above are the same as that for mode $G$. We use the mode connectivity algorithm on each of the $6$ pairs of modes including $G$ and another mode, resulting in curves $GA$, $GB$, $GC$, $GD$, $GE$, and $GF$.

Figure \ref{icml-historical} shows the validation accuracy for models on each of the $6$ connecting curves during the $20$th, $40$th, $60$th and $80$th epochs of the mode connectivity training procedure. As described in Section \ref{modeconnprocedure}, for a polychain curve $GX$ (connecting modes $G$ and $X$ using the curve described by $\theta$), model parameters $\phi_{\theta}(t)$ on the curve are given by $p_{\phi_{\theta}(t)} = 2 (t p_{\theta} + (0.5 - t) p_G) \; \text{if} \; 0 \leq t \leq 0.5$ and $p_{\phi_{\theta}(t)} = 2 ((t - 0.5)p_X + (1 - t)p_{\theta}) \; \text{if} \; 0.5 < t \leq 1$
where $p_G$, $p_{\theta}$ and $p_X$ are parameters of the models $G$, $\theta$, and $X$ respectively. Thus $\phi_{\theta}(0) = G$ and $\phi_{\theta}(1) = X$.

Within a few epochs of the curve training, for each of the 6 pairs, we can find a curve such that each point on it generalizes almost as well as models from the pair that is being connected. Thus we are able to find a high-accuracy connection between each of the 6 pairs. While connectivity for the pairs $\{G,C\}$ and $\{G,F\}$ might not be particularly surprising, one would expect the other cases to be isolated from each other and divided by high loss regions. Note that by virtue of existence of these $6$ curves, there exists a high accuracy connecting curve (albeit with multiple bends) for each of the $\Mycomb[7]{2}$ pairs of modes. We refer the reader to Figure~\ref{constellation} in the Appendix for a t-SNE plot of the modes and their connections.

Having established the high likelihood of the existence of these connecting curves, we use the curve finding procedure along with interpolating loss surface between parameters at different epochs as tools to analyze the dynamics of SGD and SGD with warm restarts (SGDR).
\vspace{-3mm}

\section{SGD with Warm Restarts}
\label{sgdrtrajectory}
\citet{loshchilov2016sgdr} introduced SGDR as an interesting modification to the cyclical LR scheme \cite{smith2017cyclical} that combines restarts with cosine annealing. 
The learning rate at the $t$-th epoch in SGDR is given by the following expression in (\ref{sgdreqn}) where $\eta_{min}$ and $\eta_{max}$ are the lower and upper bounds respectively for the LR. $T_{cur}$ represents how many epochs have been performed since the last restart and a warm restart is simulated once $T_i$ epochs are performed. Also $T_i = T_{mult} \times T_{i-1}$, meaning the period $T_{i}$ for the LR variation is increased by a factor of $T_{mult}$ after each restart. Figure \ref{sgdr_sgd_modeconnfig} (b) shows an instance of this LR schedule.
\begin{equation}
\label{sgdreqn}
\eta_{t} = \eta_{min} + \frac{1}{2}(\eta_{max} - \eta_{min})\left(1 +\cos \left(\frac{T_{cur}}{T_i}\pi\right)\right)
\end{equation}

The model returned at the end of training is the one corresponding to the iterate at the epoch just before the last restart (epoch $150$ in Figure \ref{sgdr_sgd_modeconnfig} (b)). 

While the strategy has been claimed to outperform other LR schedulers, little is known why this has been the case. One explanation that has been given in support of SGDR is that it can be useful to deal with multi-modal functions, where our iterates could get stuck in a local optimum and a restart will help them get out of it and explore another region; however, \citet{loshchilov2016sgdr} do not claim to observe any effect related to multi-modality. \citet{huang2017snapshot} propose an ensembling strategy using the set of iterates before restarts and claim that, when using the learning rate annealing cycles, the optimization path converges to and escapes from several local minima. In the next section, we try to empirically investigate if this is actually the case by interpolating the loss surface between parameters at different epochs and studying the training and validation loss for parameters on the plane passing through the two modes found by SGDR and their connectivity (plane defined by affine combinations of $w_a, w_b \text{   and   } \theta$).

\section{Loss Surface Analysis with Mode Connectivity}
\label{investigative}

We train a VGG16 network on the CIFAR10 dataset for image classification using SGD with warm restarts (SGDR). For our experiments, we choose $T_0 = 10$ epochs and $T_{mult} = 2$ (warm restarts simulated every 10 epochs and the period $T_i$ doubled at every new warm restart), $\eta_{max} = 0.05$ and $\eta_{min} = 10^{-6}$. We also perform VGG training using SGD (with momentum of $0.9$) and a step decay LR scheme (initial LR of $\eta_{0} = 0.05$, scaled by $5$ at epochs $60$ and $150$). Figure \ref{sgdr_sgd_modeconnfig} (b) shows the LR variation for these two schemes on a logarithmic scale and Figure \ref{sgdr_sgd_modeconnfig} (a) shows the validation accuracy over training epochs with these two LR schemes. 

In order to understand the loss landscape on the optimization path of SGDR, the pairs of iterates obtained just before the restarts 
$\{w_{30},w_{70}\},\{w_{70},w_{150}\}$ and $\{w_{30},w_{150}\}$ are given as inputs to the mode connectivity algorithm, where $w_n$ is the model corresponding to parameters at the $n$-th epoch of the SGDR training. Figure \ref{sgdr_sgd_modeconnfig} (c) shows the training loss for models along the line segment joining these pairs and those on the curve found through mode connectivity. For the baseline case, we connect the iterates around the epochs when we decrease our LR in the step decay LR scheme. Thus we chose $\{w_{55},w_{65}\},\{w_{145},w_{165}\}$ and $\{w_{55},w_{165}\}$ as input pairs to the mode connectivity algorithm where now $w_n$ is the model corresponding to parameters at the $n$-th epoch of SGD with the step decay LR scheme.  Figure \ref{sgdr_sgd_modeconnfig} (d) shows the training loss for models along the line segments joining these pairs and the curves found through mode connectivity.

From Figure \ref{sgdr_sgd_modeconnfig} (c), it is clear that for the pairs $\{w_{30},w_{150}\}$ and $\{w_{70},w_{150}\}$ the training loss for points on the line segment is much higher than the endpoints suggesting that SGDR indeed finds paths that move over a barrier\footnote{a path is said to have moved over or crossed a barrier between epoch $m$ and $n$ ($n > m$) if $\exists$ $w_t \in \{\lambda w_m + (1 - \lambda) w_n | \lambda \in  [0,1] \}$ such that $\mathcal{L}(w_t) > \max \{\mathcal{L}(w_m),\mathcal{L}(w_n) \} $} in the training loss landscape. In contrast, for SGD (without restarts) in Figure \ref{sgdr_sgd_modeconnfig} (d) none of the three pairs show evidence of having a training loss barrier on the line segment joining them. Instead there seems to be an almost linear decrease of training loss along the direction of these line segments, suggesting that SGD's trajectory is quite different from SGDR's. We present additional experiments, including results for other metrics, in Appendix~\ref{section:additionalexperiments}.

\begin{figure}[ht]
\begin{center}
\centerline{\includegraphics[width=\columnwidth]{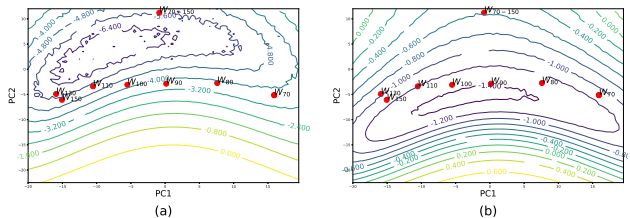}}
\caption{(a) Training loss surface (log scale) and (b) validation loss surface (log scale) for points (models) on the plane defined by $\{W_{70},W_{150},W_{70-150}\}$ including projections of the SGDR iterates on this plane}. 
\label{train_contour}
\end{center}
\vskip -0.2in
\end{figure}

To understand the SGDR trajectory more concretely, we evaluate the intermediate iterates on the plane in the $D$-dimensional space defined by the three points: $w_{70}$, $w_{150}$ and $w_{70-150}$, where $w_{70-150}$ is the $\theta$ that defines the high accuracy connection for the pair $\{w_{70},w_{150}\}$. This $2$-d plane in the $D$-dimensional space consists of all the affine combinations of $w_{70}$, $w_{150}$ and $w_{70-150}$. Figure \ref{train_contour} (a) and \ref{train_contour} (b) show the training and validation loss surface for points in this subspace, respectively. Note that the intermediate iterates do not necessarily lie in this plane and thus need to be projected. Hence, one cannot tell the value of loss at the actual iterates from their representation in Figure \ref{train_contour}. We refer the reader to Appendix~\ref{section:projection} for additional details regarding the projection process and Appendix~\ref{section:sgdr30to70} for analogous results with $w_{30}$ and $w_{70}$. 

Figure \ref{train_contour} (a) suggests that SGDR helps the iterates to converge to a different region although neither of $w_{70}$ or $w_{150}$ are technically a local minimum, nor do they appear to be lying in different \textit{basins}, hinting that \citet{huang2017snapshot}'s claims about SGDR converging to and escaping from local minima might be an oversimplification. 

Another insight we can draw from Figure \ref{train_contour} (a) is that the path found by mode connectivity corresponds to lower training loss than the loss at the iterates that SGDR converges to ($\mathcal{L}(w_{150}) > \mathcal{L}(w_{70-150})$). However, Figure \ref{train_contour} (b) shows that models on this curve seem to overfit and not generalize as well as the iterates $w_{70}$ and $w_{150}$ which stands as further evidence that SGD's stochasticity helps generalization. This observation is also consistent with what we see in Figure \ref{icml-historical}. Thus, gathering models from this connecting curve might seem as a novel and computationally cheap way of creating ensembles, this generalization gap alludes to one limitation in doing so; \citet{garipov2018loss} point to other shortcomings of curve ensembling in their original work. 

In Figure \ref{train_contour}, the region of the plane under consideration, between the iterates $w_{70}$ and $w_{150}$, corresponds to higher training loss but lower validation loss than the two iterates. This hints at a reason why averaging iterates to improve generalization using cyclic or constant learning rates \cite{izmailov2018averaging} has been found to work reasonably well.   

\section{Conclusion}

We revisited the recently proposed mode connectivity procedure, and explored its limits by using it to find the desired connection between models trained with different training schemes and initializations. Remarkably, we found curves with reasonably high accuracy for mode pairs that we considered. These results are indicative of the connectedness of deep learning loss surface minima. Given this resiliency, we use the framework to inspect the claims made to explain the effectiveness of restarts and cosine annealing in SGDR, and studied the SGDR trajectory using the subspace defined by the mode connections. We found that although SGDR tends to move over barriers, claims about SGDR converging to and escaping multiple local minima are not substantied by our experiments. Our work establishes the wide generality of the mode connectivity framework, and encourages use of it as a tool for understanding not just the training landscape but also the training trajectories.

\nocite{langley00}


\bibliography{example_paper}
\bibliographystyle{icml2018}
\clearpage
\newpage

\appendix

\section{Additional Results}
\subsection{t-SNE visualization for the 7 modes}
\begin{figure}[ht]
\vskip 0.2in
\begin{center}
\centerline{\includegraphics[scale=0.35]{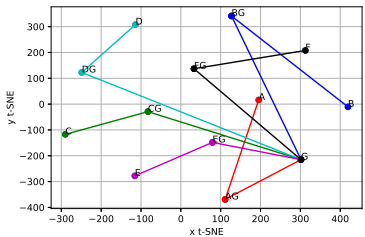}}
\caption{Representing the modes and their connecting point using t-SNE}
\label{constellation}
\end{center}
\vskip -0.2in
\end{figure}

We use t-SNE \cite{maaten2008visualizing} to visualize these $7$ modes and the $\theta$ points that define the connectivity for the $6$ pairs presented in Section \ref{resilience}, in a $2$-dimensional plot in Figure \ref{constellation}. Since t-SNE is known to map only local information correctly and not preserve global distances, we caution the reader about the limited interpretability of this visualization, it is presented simply to establish the notion of connected modes. 

It is interesting to note that although neural networks performance are quite resilient to changes in weight (and bias) parameters as suggested by the mode connectivity phenomenon, that is not the case when it comes to perturbations to the input of the network \cite{szegedy2013intriguing}.
\subsection{Projecting iterates}
\label{section:projection}
The $W_n$ in Figure \ref{train_contour}(a) and \ref{train_contour}(b) is equivalent to $$W_n = P_c(w_n) = {\lambda^\star}^\top \begin{bmatrix}
w_{70} \\
w_{150} \\
\theta
\end{bmatrix} $$
\\
$$
\text{where} \; {\lambda^\star} = \text{argmin}_{\lambda \in \mathbb{R}^3} ||\lambda^\top \begin{bmatrix}
           w_{70} \\
           w_{150} \\
           \theta
         \end{bmatrix} - w_n ||_2^2 $$ 
meaning it is the point on the plane (linear combination of $w_{70}, w_{150} \; \text{and} \; \theta$) with the least $l$-2 distance from the original point (iterate in this case). 

\subsection{Connecting modes $W_{30}$ and $W_{70}$ from SGDR}
\label{section:sgdr30to70}
In Section \ref{sgdrtrajectory}, we present some experiments and make observations on the trajectory of SGDR by using the plane defined by the points $w_{70}$,$w_{150}$ and $w_{70-150}$. Here we provide the loss surface plots for another plane defined by SGDR's iterates $\{w_{30},w_{70},w_{30-70}\}$ to ensure the reader that the observations made are general enough.

\begin{figure}[ht]
\begin{center}
\centerline{\includegraphics[scale=0.275]{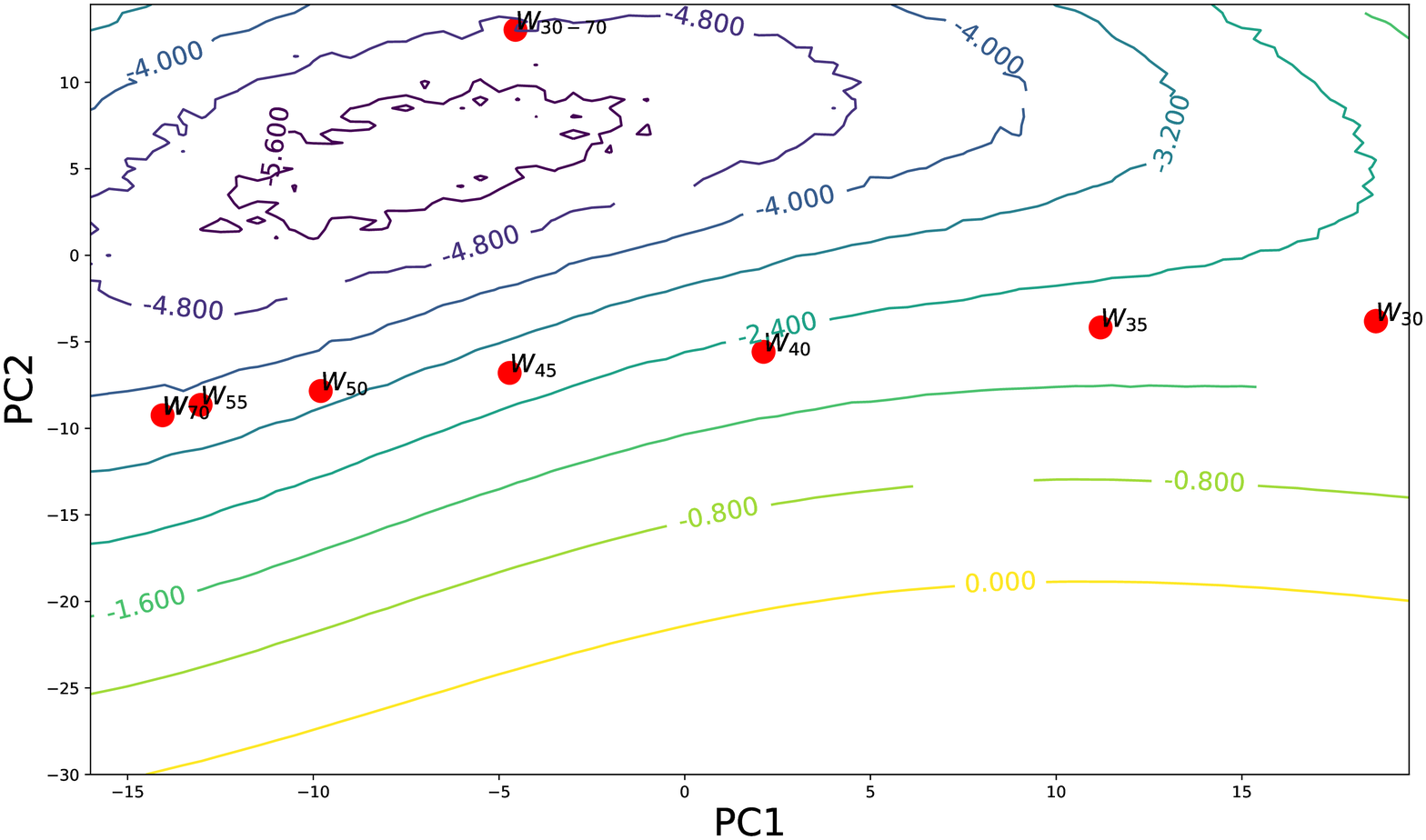}}
\caption{Training Loss Surface (log scale)}
\label{app_train_contour}
\end{center}
\vskip -0.2in
\end{figure}

\begin{figure}[ht]
\begin{center}
\centerline{\includegraphics[scale=0.275]{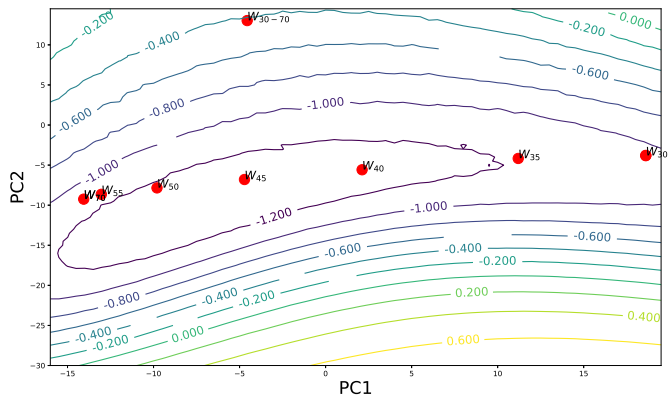}}
\caption{Validation Loss Surface (log scale)}
\label{app_val_contour}
\end{center}
\vskip -0.2in
\end{figure}

\subsection{Additional Experiments}
\label{section:additionalexperiments}
\begin{figure*}[ht]
\centering{\includegraphics[scale = 0.35]{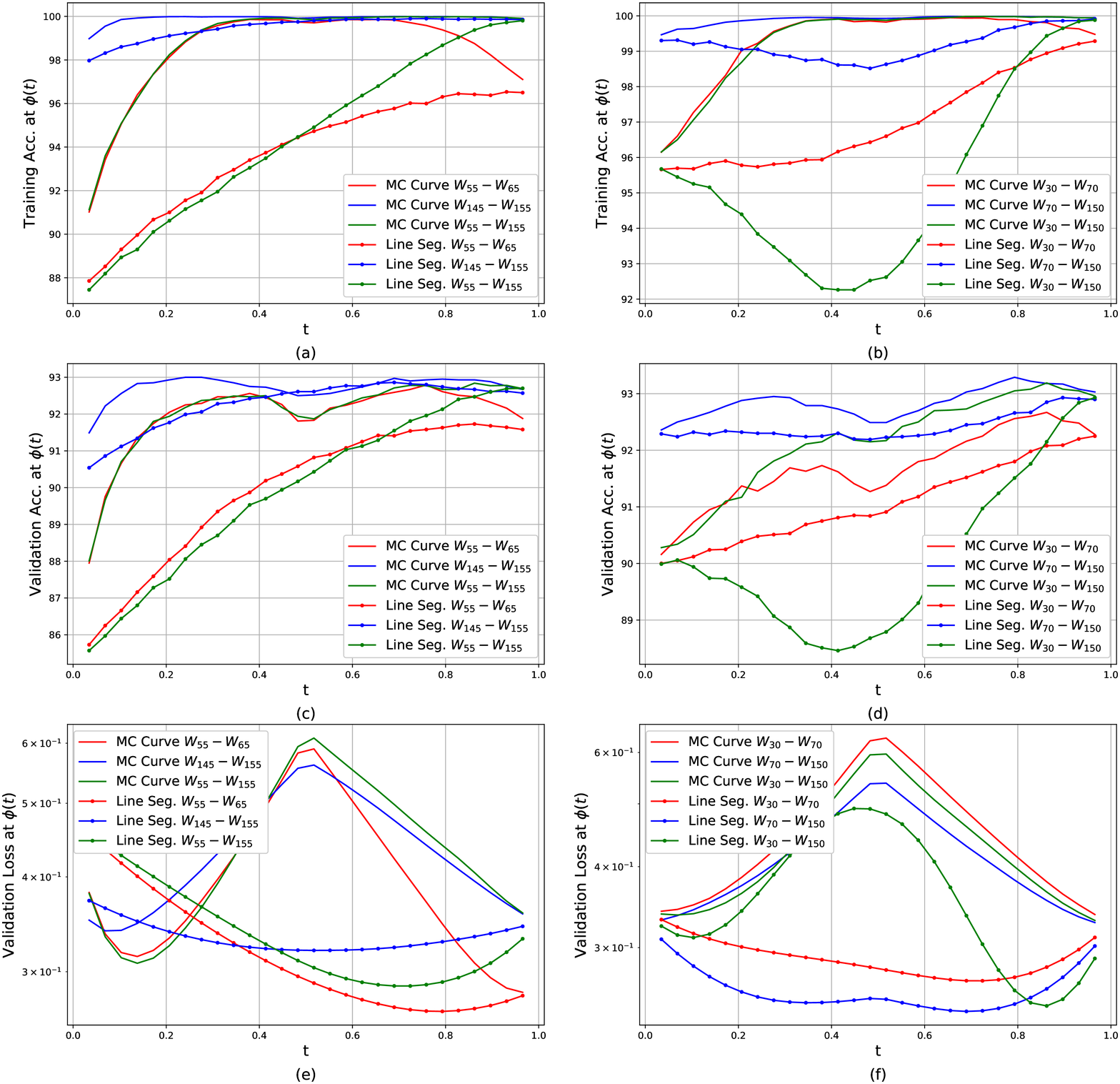}}
\caption{\textbf{Left} Column: Connecting iterates from SGD with step-decay LR scheme \textbf{Right} Column: Connecting iterates from SGDR \\ \textbf{Top} Row: Training Accuracy on the curve found through Mode Connectivity (MC Curve) and on the line segment (Line Seg.) joining iterates from SGDR and SGD. \textbf{Middle} row: Validation Accuracy on the curve found through Mode Connectivity (MC Curve) and on the line segment (Line Seg.) joining iterates from SGDR and SGD.  \textbf{Bottom} row Validation Loss on the curve found through Mode Connectivity (MC Curve) and on the line segment (Line Seg.) joining iterates from SGDR and SGD.
}
\label{addnalsec2plot}
\vskip -0.2in
\end{figure*}

For completeness, we present Figure \ref{addnalsec2plot} plotting the remaining quantities - namely Training Accuracy, Validation Accuracy and Validation Loss over the connecting curve from Section \ref{investigative}, Figure \ref{sgdr_sgd_modeconnfig} for the pair of iterates obtained using SGD and SGDR.

\begin{figure}[ht]
\begin{center}
\centering{\includegraphics[scale = 0.3]{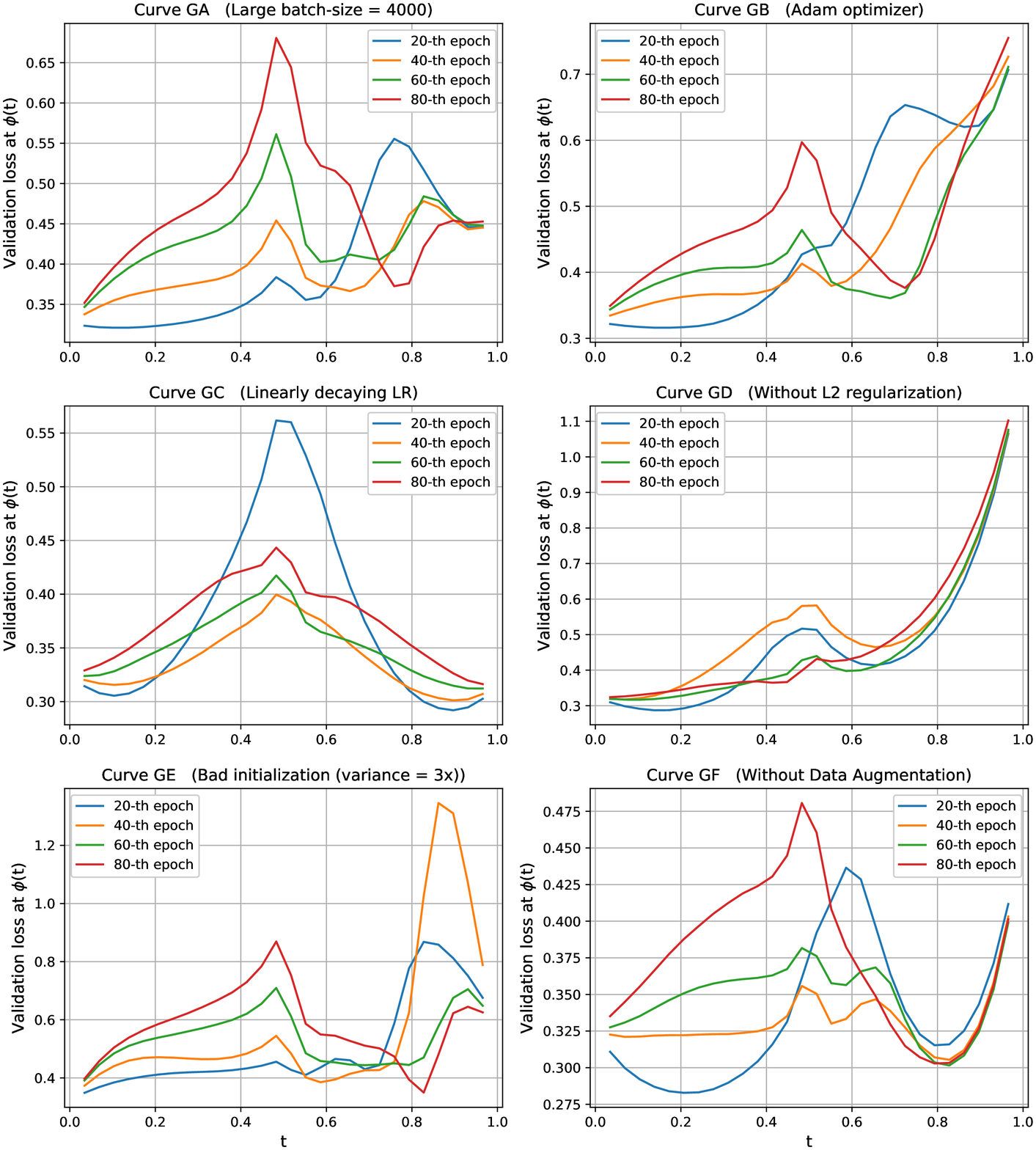}}
\caption{Validation Loss corresponding to models on the following 6 different curves - curve $GA$ represents curve connecting mode $G$ (one found with all the default hyperparameters) and mode $A$ (found using large batch size), similarly, curve $GB$ connects mode $G$ and mode $B$ (found using Adam), curve $GC$ connects to mode $C$ (found using linearly decaying LR), curve $GD$ to mode $D$ (found with much less L2 regularization), curve $GE$ to mode $E$ (found using a poor initialization), and curve $GF$ to mode $F$ (found without using data augmentation).}
\label{icml-historical-testloss}
\end{center}
\vskip -0.2in
\end{figure}

\begin{figure}[ht]
\begin{center}
\centering{\includegraphics[scale = 0.3]{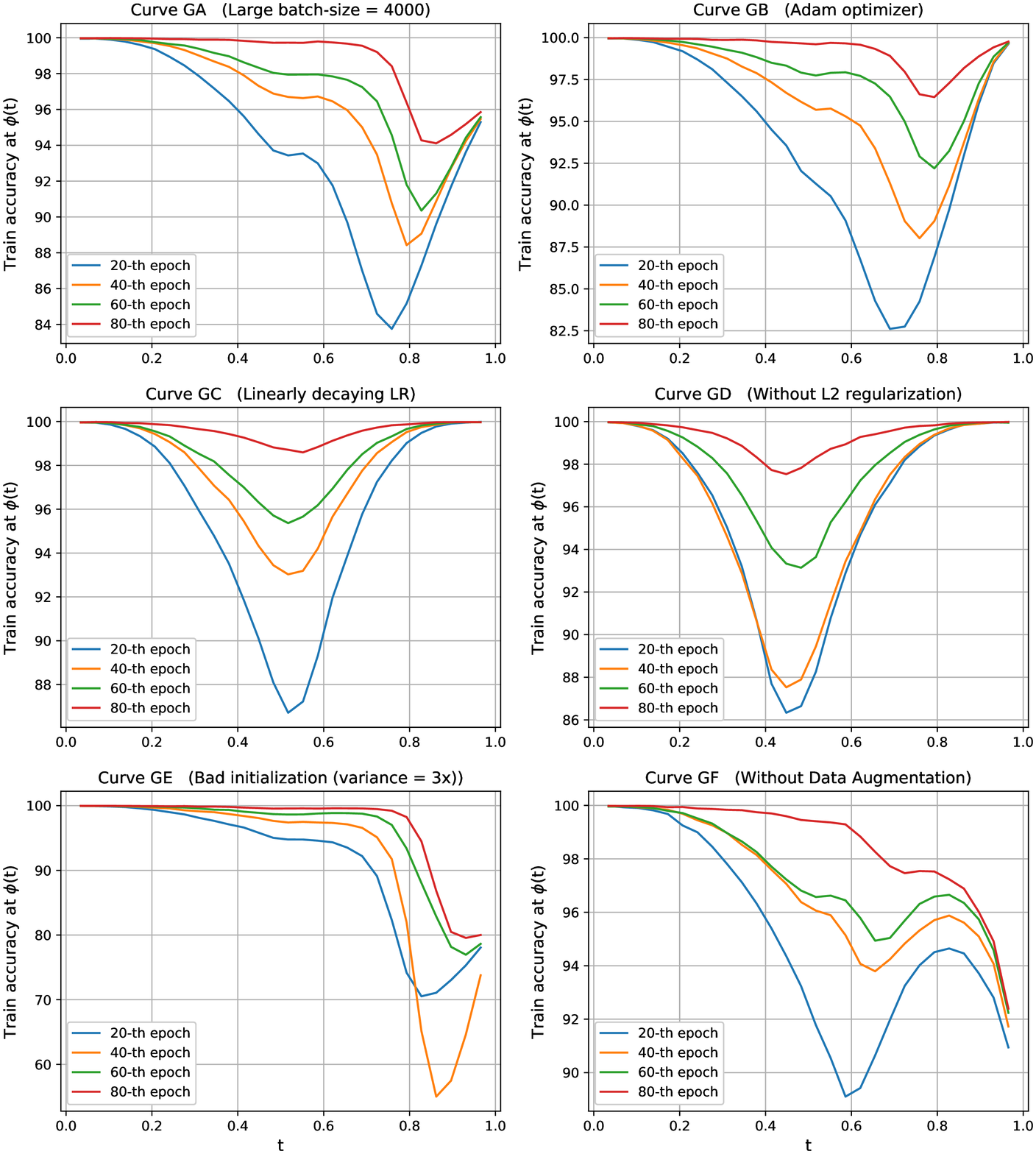}}
\caption{Training accuracy corresponding to models on the following 6 different curves - curve $GA$ represents curve connecting mode $G$ (one found with all the default hyperparameters) and mode $A$ (found using large batch size), similarly, curve $GB$ connects mode $G$ and mode $B$ (found using Adam), curve $GC$ connects to mode $C$ (found using linearly decaying LR), curve $GD$ to mode $D$ (found with much less L2 regularization), curve $GE$ to mode $E$ (found using a poor initialization), and curve $GF$ to mode $F$ (found without using data augmentation).}
\label{icml-historical-trainacc}
\end{center}
\vskip -0.2in
\end{figure}

\begin{figure}[ht]
\begin{center}
\centering{\includegraphics[scale = 0.3]{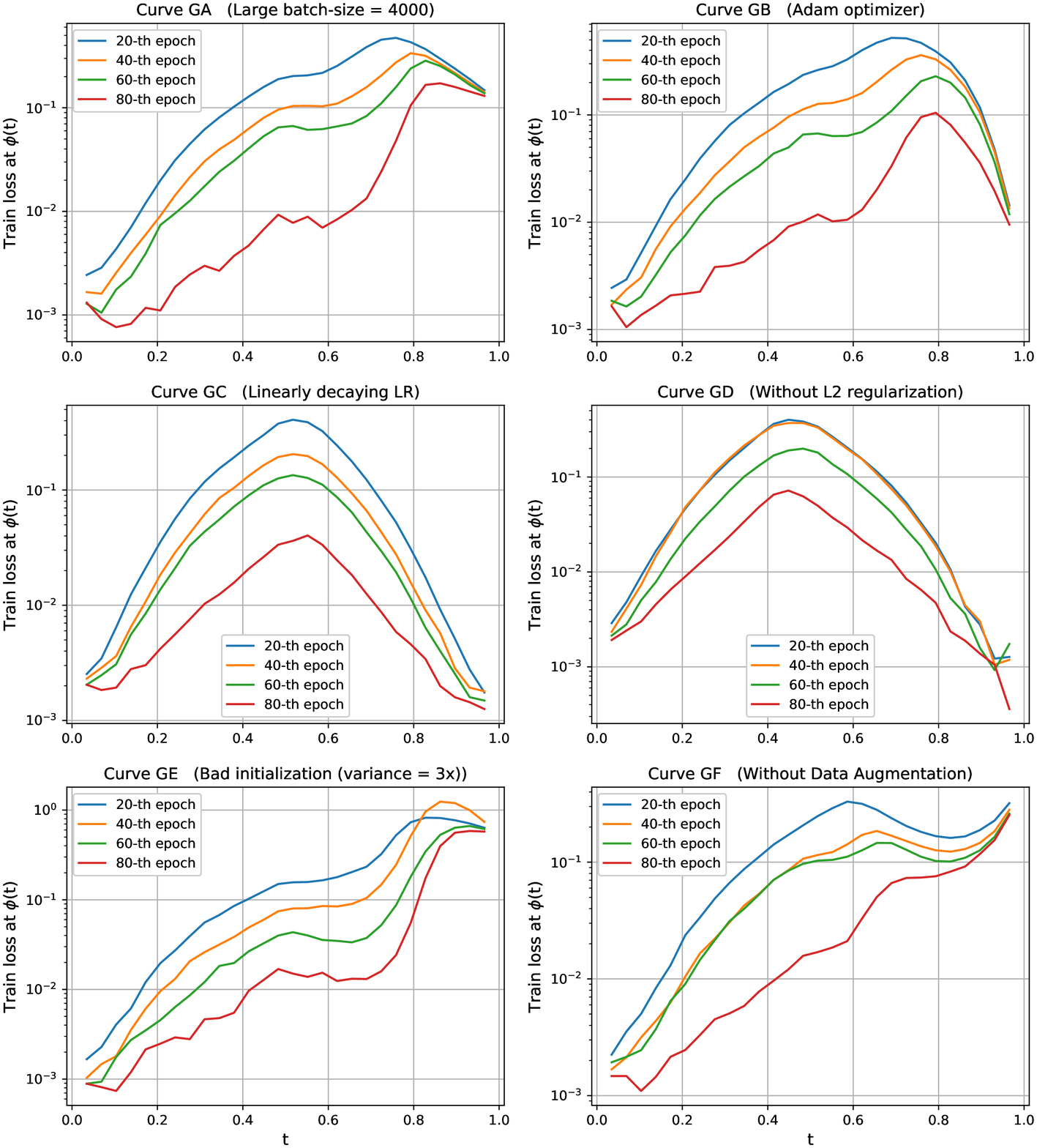}}
\caption{Training Loss corresponding to models on the following 6 different curves - curve $GA$ represents curve connecting mode $G$ (one found with all the default hyperparameters) and mode $A$ (found using large batch size), similarly, curve $GB$ connects mode $G$ and mode $B$ (found using Adam), curve $GC$ connects to mode $C$ (found using linearly decaying LR), curve $GD$ to mode $D$ (found with much less L2 regularization), curve $GE$ to mode $E$ (found using a poor initialization), and curve $GF$ to mode $F$ (found without using data augmentation).}
\label{icml-historical-trainloss}
\end{center}
\vskip -0.2in
\end{figure}

Figures \ref{icml-historical-testloss}, \ref{icml-historical-trainacc} and \ref{icml-historical-trainloss} show the Validation Loss, Training Accuracy and Training Loss respectively for the curves joining the $6$ pairs discussed in Section \ref{resilience}. These results too, confirm the overfitting or poor generalization tendency of models on the curve.

\end{document}